\documentclass[runningheads]{llncs}
\usepackage{caption}
\usepackage[T1]{fontenc}
\usepackage{amsmath,amssymb,amsfonts}
\usepackage{algorithmic}
\usepackage{algorithm,algorithmic}
\usepackage{textcomp}
\usepackage{subcaption}

\usepackage{booktabs}
\usepackage{multirow}
\usepackage{xcolor}

\usepackage{bm}
\usepackage{graphicx,verbatim}
\begin{document}
\title{Cross-Modal Prototype Allocation: Unsupervised Slide Representation Learning via Patch-Text Contrast in Computational Pathology}
%

\author{
Yuxuan Chen\inst{1}\textsuperscript{*} \and
Jiawen Li\inst{1}\textsuperscript{*} \and
Jiali Hu\inst{1} \and
Xitong Ling\inst{1} \and
Tian Guan\inst{1}\textsuperscript{\textdagger} \and
Anjia Han\inst{2}\textsuperscript{\textdagger} \and
Yonghong He\inst{1,3}\textsuperscript{\textdagger}
}

\begingroup
\renewcommand\thefootnote{\relax}
\footnotetext{* Contributed equally.}
\footnotetext{\textdagger\ Corresponding author.}
\endgroup

%
%
\institute{
Shenzhen International Graduate School, Tsinghua University, China 
\email{\{chenyx23, jw-li24\}@mails.tsinghua.edu.cn} \\
\email{heyh@sz.tsinghua.edu.cn}
\and
Department of Pathology, The First Affiliated Hospital of Sun Yat-sen University, China \\
\email{hananjia@mail.sysu.edu.cn}
}

\maketitle              
\begin{abstract}

With the rapid advancement of pathology foundation models (FMs), the representation learning of whole slide images (WSIs) attracts increasing attention. Existing studies develop high-quality patch feature extractors and employ carefully designed aggregation schemes to derive slide-level representations. However, mainstream weakly supervised slide representation learning methods, primarily based on multiple instance learning (MIL), are tailored to specific downstream tasks, which limits their generalizability. To address this issue, some studies explore unsupervised slide representation learning. However, these approaches focus solely on the visual modality of patches, neglecting the rich semantic information embedded in textual data. In this work, we propose ProAlign, a cross-modal unsupervised slide representation learning framework. Specifically, we leverage a large language model (LLM) to generate descriptive text for the prototype types present in a WSI, introducing patch-text contrast to construct initial prototype embeddings. Furthermore, we propose a parameter-free attention aggregation strategy that utilizes the similarity between patches and these prototypes to form unsupervised slide embeddings applicable to a wide range of downstream tasks. Extensive experiments on four public datasets show that ProAlign outperforms existing unsupervised frameworks and achieves performance comparable to some weakly supervised models.


\end{abstract}
\section{Introduction}

With the rise of computational pathology, the analysis of WSIs has garnered increasing attention, with slide representation learning being a key focus. The current mainstream approach to slide representation learning is based on MIL for weakly supervised learning. In this paradigm, a WSI is first divided into multiple patches, and the initial embeddings of these patches are extracted using a pre-trained encoder. Then, the patch features are aggregated to form slide-level embeddings using attention mechanisms \cite{ABMIL,CLAM,agent}, graph convolutions \cite{Patch-GCN,WiKG,DyHG}, transformers \cite{TransMIL,RetMIL}, and other techniques \cite{FRMIL,DSMIL,rrtmil,shapley,mergeup}. The slide representations obtained by these methods rely on the supervision of slide-level labels in the learning process and have been widely proven to be effective in specific tasks.

Despite the success of MIL-based weakly supervised learning methods, several challenges remain. First, these methods rely on slide-level labels to supervise the learning of slide representations, and the quality of slide representation learning is heavily dependent on the accuracy of slide labels. However, label noise in pathology datasets is almost inevitable. \cite{intro1} shows that the inter-observer agreement among different doctors on the same breast biopsy dataset is only 75\%. The unavoidable noisy labels can significantly impact the quality of slide representation learning. Second, the embeddings learned by these methods are often task-specific and exhibit task dependence. As a result, their performance typically degrades when transferred to other datasets, resulting in poor generalization. Lastly, to address the limitation of traditional MIL methods in capturing rich contextual information between patches, the model structures in these methods are often complex (e.g., transformers, GCNs). While effective, these methods are not optimal in terms of efficiency and are less friendly for clinical application. 

In light of these issues, unsupervised slide representation learning, with its advantages of no label dependency, no downstream task dependence, and simpler model structures, has garnered increasing attention. Recent works on unsupervised slide representation learning \cite{H2T,ProtoCount,OT,PANTHER} have demonstrated promising performance. However, they primarily leverage image-modality information, neglecting the rich semantic information contained in the text modality.

Based on these discoveries, we present ProAlign, a cross-modal slide representation learning framework based on the hypothesis that a WSI can be represented by multiple prototypes. First, we prompt an LLM to obtain descriptive texts for each prototype category. Then, we use visual-language (V-L) pathology FMs, such as CONCH \cite{conch} and PLIP \cite{plip}, to extract the embeddings of the patches and prototype descriptions. For the initial prototype embedding construction, we propose a patch-text contrast method, where prototypes are formed by clustering patches based on the similarity between patch embeddings and prototype description embeddings. Next, based on the initial prototype embeddings, we propose a parameter-free attention aggregation mechanism to refine prototype embeddings that are specific to each WSI. Finally, we concatenate all the refined prototype embeddings to obtain the final slide embeddings, which are used for downstream tasks. We conduct extensive experiments on four publicly available datasets. The results show that ProAlign outperforms existing unsupervised baselines and achieves performance comparable to several weakly supervised baselines.

\begin{figure}[ht]
  \centering
  \includegraphics[width=0.98\linewidth]{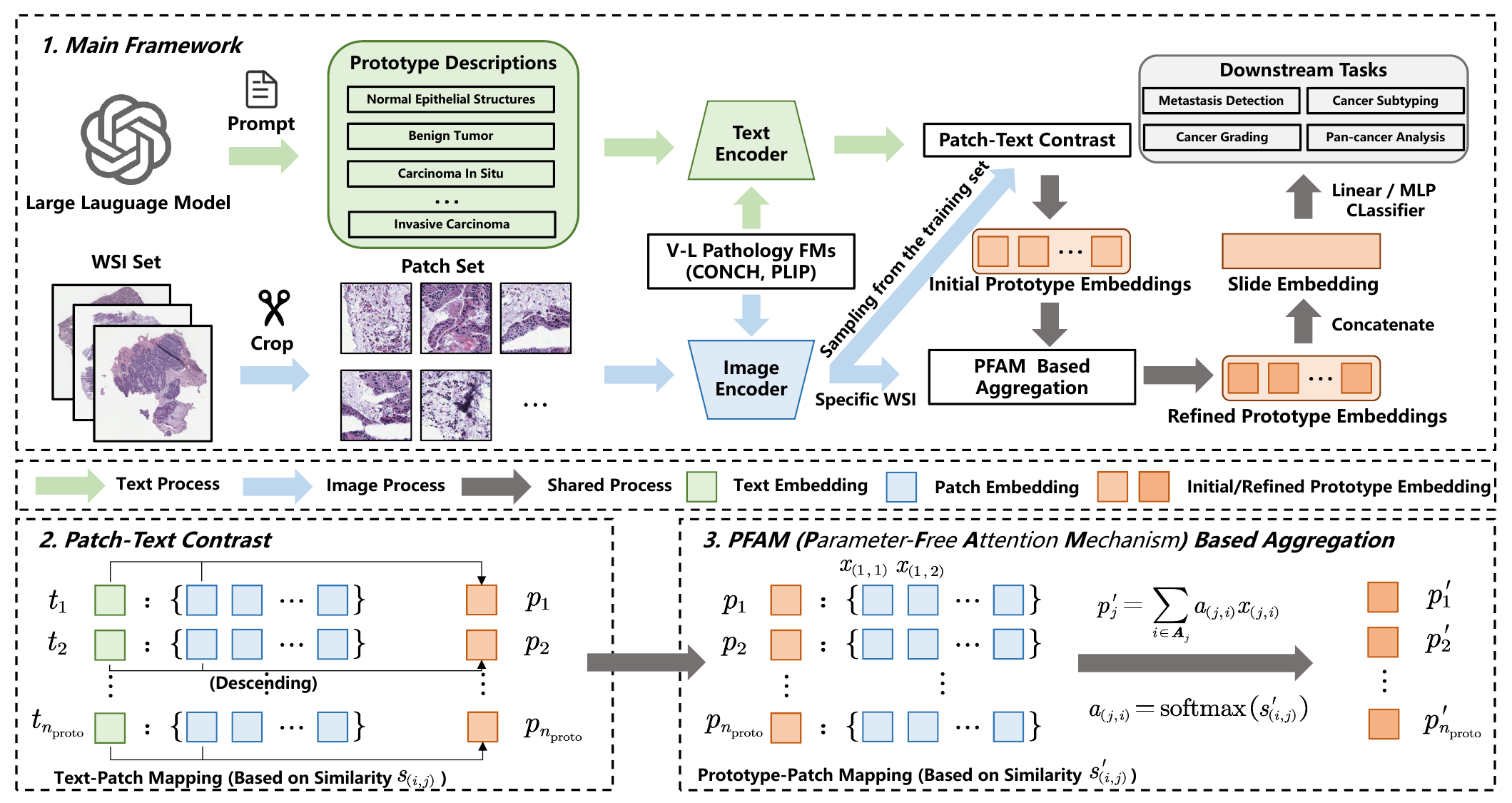}
   \caption{The framework of ProAlign: WSIs are cropped into patches and protptype descriptions are obtained through prompting LLM. Text and image encoders are used to extract features. Next, patch-text contrast is exploited to obtain initial prototype embeddings and PFAM based aggregation is used to refine prototype embeddings for specific WSI. Finally, the refined prototype embeddings are concatenated to form slide embeddings for downstream tasks.}
   \label{fig:main_fig}
\end{figure}

\section{Methodology}

In this section, we introduce the workflow of ProAlign, including the construction of initial prototype embeddings, the construction of prototype embeddings and slide emebeddings for specific WSIs and downstream evaluation. The framework is shown in Fig. \ref{fig:main_fig}. 

\subsection{Patch-text contrast}

Given a dataset containing \(N\) WSIs, we first partition the dataset into training, validation, and test sets, which contain $N_{train}$, $N_{val}$ and $N_{test}$ WSIs, respectively. For each WSI, we segment it into patches and then extract the initial patch embeddings using the image encoder of a V-L pathology FM. Following \cite{PANTHER}, we set the number of prototypes at \(n_{\text{proto}}\) and the number of patches required for each prototype to \(n_{\text{patch\_per\_proto}}\), resulting in a total of \(n_{\text{total}} = n_{\text{proto}} \times n_{\text{patch\_per\_proto}}\) patches. We then randomly select \(\frac{n_{\text{total}}}{N_{\text{train}}}\) patches from each WSI in the training set to construct the patch set to learn initial prototype embeddings, and the corresponding embedding matrix is denoted as \(\bm X_{\text{proto}} \in \mathbb{R}^{n_{\text{total}}\times d}\).

Existing work \cite{PANTHER} typically constructs the initial prototype embeddings by applying k-means clustering to \(\bm X_{\text{proto}}\), which is computationally expensive and fails to take advantage of the rich semantic information contained in the textual modality. Instead, we propose patch-text contrast to construct initial prototype embeddings. Specifically, we prompt an LLM to obtain the description of each prototype. Subsequently, we use the text encoder of a V-L pathology FM to extract the embeddings of these prototype descriptions, forming the set of prototype text features \(\bm T_{\text{proto}} = \{\bm t_1, \bm t_2, \cdots, \bm t_{n_{\text{proto}}}\}\), where \(\bm t_i \in \mathbb{R}^{1 \times d}\).

We then divide the patches into different prototypes by patch-text contrast. Specifically, we first calculate the similarity matrix $S$ between patches and texts:

\begin{equation}
\bm S = \bm X_{\text{proto}}\bm T_{\text{proto}}^{\top},
\end{equation}
where $\bm S\in \mathbb{R}^{n_{total}\times n_{proto}}$, $\bm s_{(i,j)} \in \bm S$ denotes the similarity between patch $i$ and prototype $j$.

The initial embedding of prototype \(j\) is composed of the sum of its text embedding and the embedding of the patch with the highest similarity:
\begin{equation}
\bm p_j = \bm t_j + \bm x_h,
\end{equation}
where \(\bm x_h = \bm X_{\text{Proto}}(i^*)\) with \(i^* = \arg\max_{i \in \{1,2,\ldots,n_{\text{total}}\}} \bm s_{(i,j)}\).

Then we obtain the initial embedding matrix of prototypes denoted as $\bm P=\{\bm p_1, \bm p_2, \cdots, \bm p_{n_{proto}}\}$.

\subsection{Parameter-free attention mechanism}
To construct prototype embeddings for specific WSIs, we introduce a parameter-free attention mechanism (PFAM) based on the initial embedding matrix of prototypes $\bm P$. Specifically, given a WSI with its patch embedding matrix $\bm X\in \mathbb{R}^{n \times d}$, we calculate the similarity matrix $S^{\prime}$ between patches and prototypes:
\begin{equation}
\bm S^{\prime} = \bm X\bm P^{\top},
\end{equation}
where $\bm S^{\prime}\in \mathbb{R}^{n\times n_{proto}}$, $\bm s^{\prime}_{(i,j)} \in \bm S^{\prime}$ denotes the similarity between patch $i$ and prototype $j$.

Each patch is assigned to the prototype with the highest similarity. Then for each prototype, PFAM is performed to aggregate the embeddings of patches that belong to it:
\begin{equation}
\bm p^{\prime}_{j} = \sum_{i\in \bm A_j}\bm a_{(j,i)}\bm x_{(j,i)}, ~~~\bm a_{(j,i)} = \frac{\exp(\bm s'_{(i,j)})}{\sum_{l\in \bm A_j} \exp(\bm s'_{(l,j)})},
\end{equation}
where $\bm p^{\prime}_{j} \in \mathbb{R}^{1\times d}$ denotes the refined embedding of prototype $j$, $\bm A_j$ is the set of indexs of all patches assigned to prototype $j$, $\bm x_{(j,i)}$ denotes the embedding of patch $i$ that belongs to prototype $j$, and $\bm a_{(j,i)}$ is the attention score of patch $i$, which is obtain in a parameter-free manner via the softmax function.



It is worth mentioning that for prototypes that are not assigned to patches, we directly use their initial embeddings as the refined embeddings.

Then we obtain the refined embedding matrix of prototypes for a specific WSI denoted as $\bm P^{\prime}=\{\bm p^{\prime}_1, \bm p^{\prime}_2, \cdots, \bm p^{\prime}_{n_{proto}}\}$.

\subsection{Downstream evaluation}

For a given WSI, we concatenate the refined embeddings obtained in the previous step as WSI level embedding, denoted as:
\begin{equation}
\bm X^{\prime}=[\bm p^{\prime}_1,\bm p^{\prime}_2, \cdots, \bm p^{\prime}_{n_{proto}}]
\end{equation}
Then we refer to \cite{PANTHER} to use linear layers (Lin.) or multi-layer perceptrons (MLP) as predictor $f(\cdot)$ for various downstream tasks.



\section{Experiments}
\subsection{Settings}


To evaluate the performance of ProAlign, we use four publicly available datasets, each corresponding to a different downstream task: \textbf{CAMELYON+ \cite{camelyon+}}, which integrates CAMELYON16 and CAMELYON17, corresponding to the downstream task of metastatic cancer analysis; \textbf{TCGA-NSCLC\cite{TCGA}}, which focuses on non-small cell lung cancer and is structured as a binary classification task, supporting a subtyping downstream task; \textbf{PANDA\cite{panda}}, dedicated to prostate cancer grading and including six distinct grades, corresponding to the Gleason scoring system; and \textbf{CPTAC\cite{cptac}}, a pan-cancer dataset encompassing 11 diverse cancer types, corresponding to the downstream task of pan-cancer analysis. For each dataset, we split the data of each dataset into training, validation, and test sets at a ratio of 6:2:2.


In the preprocessing stage, we crop each WSI into 256x256 patches at 20x magnification, and then extract features using FMs (e.g., CONCH \cite{conch} and PLIP \cite{plip}). We set the number of prototypes to 16, with each prototype requiring $10^5$ patches for training, like \cite{PANTHER}, and prompt a LLM with the following query: \textit{"Please divide a WSI into 16 prototype regions, ensuring that patches at 20x magnification are assigned to one of these prototype regions. Provide the name and description for each prototype region."}. This generates a description for each prototype region. In terms of experimental setup, we use AdamW as optimizer with a decay rate of $10^{-5}$ and a learning rate of $10^{-4}$. For all unsupervised models, we use balanced accuracy (B acc) and weighted F1 score (F1) as evaluation metrics to assess the model's performance. All experiments are conducted on a single NVIDIA A6000 GPU.

\subsection{Comparison Results}
\begin{table*}[htbp]
    \centering
    \caption{Comparison results on different classification tasks based on CONCH. The best results of supervised baselines are in red. The best results of unsupervised models are in bold, and the second best ones are in blue.}
    \label{conch}
    \resizebox{\linewidth}{!}{
    \begin{tabular}{clcccccccccccccc}
        \toprule
        & \multirow{2}{*}{\textbf{Method}} & \multicolumn{2}{c}{\textbf{CAMELYON+}} & & \multicolumn{2}{c}{\textbf{TCGA-NSCLC}} & & \multicolumn{2}{c}{\textbf{PANDA}} & & \multicolumn{2}{c}{\textbf{CPTAC}} \\
        \cmidrule{3-4} \cmidrule{6-7} \cmidrule{9-10} \cmidrule{12-13} 
        & & \textbf{B acc} & \textbf{F1} & & \textbf{B acc} & \textbf{F1} & & \textbf{B acc} & \textbf{F1} & &  \textbf{B acc} & \textbf{F1}\\
        \midrule
        \multirow{8}{*}{\rotatebox{90}{\textbf{Supervised.}}} 
        & MAXMIL & $49.17_{0.27}$ & $74.31_{0.28}$ & & $84.97_{0.48}$ & $84.97_{0.48}$ & & $48.26_{0.30}$ & $53.88_{0.27}$ & & $86.61_{0.46}$ & $87.08_{0.37}$ \\
        & MEANMIL & $37.96_{0.22}$ & $66.07_{0.20}$ & & $83.07_{0.19}$ & $83.07_{0.19}$ & & $46.65_{0.12}$ & $51.77_{0.12}$ & & $87.66_{0.14}$ & $87.80_{0.13}$ \\
        & ABMIL \cite{ABMIL} & $55.94_{3.61}$ & $80.56_{3.46}$ & & $86.24_{0.45}$ & $86.24_{0.45}$ & & $53.43_{0.13}$ & $59.38_{0.11}$ & & $87.92_{0.23}$ & $88.22_{0.17}$\\
        & DSMIL \cite{DSMIL} & $43.36_{1.46}$ & $70.40_{1.09}$ & & $82.89_{0.55}$ & $82.89_{0.55} $ & & $50.09_{0.31}$ & $55.47_{0.30}$ & & $88.66_{0.22}$ & $88.89_{0.16}$ \\
        & TransMIL \cite{TransMIL} & $\textcolor{red}{65.55_{1.38}}$ & $\textcolor{red}{87.33_{0.87}}$ & & $87.02_{1.10}$ & $87.01_{1.09}$ & & $54.78_{0.41}$ & $60.35_{0.57}$ & & $92.32_{0.42}$ & $91.92_{0.34} $ \\
        & RRTMIL \cite{rrtmil} & $65.36_{3.51}$ & $85.73_{1.96}$ & & $86.89_{1.09}$ & $86.88_{1.09}$ & & $\textcolor{red}{55.59_{1.16}}$ & $\textcolor{red}{61.29_{1.18}}$ & & $\textcolor{red}{92.86_{0.90}}$ & $\textcolor{red}{92.88_{0.59}}$ \\
        & WiKG \cite{WiKG} & $56.34_{0.62}$ & $81.35_{0.68}$ & & $\textcolor{red}{87.45_{0.23}}$ & $\textcolor{red}{87.45_{0.23}} $ & & $55.21_{0.54}$ & $60.13_{0.46}$ & & $91.95_{0.70}$ & $92.04_{0.36}$ \\
        & FRMIL \cite{FRMIL} & $51.90_{5.36}$ & $77.25_{4.66}$ & & $84.17_{1.61}$ & $84.21_{1.72}$ & & $53.04_{0.48}$ & $58.89_{0.40}$ & & $86.71_{0.67}$ & $87.24_{0.67}$ \\
        \midrule
        \multirow{5}{*}{\rotatebox{90}{\textbf{Unsup.}}} 
        & H2T \cite{H2T} & $25.00_{0.00}$ & $50.51_{0.00}$ & & $84.04_{0.13}$ & $84.04_{0.13}$ & & $42.85_{0.08}$ & $47.32_{0.07}$ & & $78.79_{0.24}$ & $80.03_{0.18}$ \\
        & ProtoCount \cite{ProtoCount} & $25.04_{5.18}$ & $44.72_{7.72}$ & & $59.94_{5.91}$ & $58.04_{5.93}$ & & $29.88_{3.48}$ & $33.19_{3.60}$ & & $35.02_{11.87}$ & $36.12_{7.15}$ \\
        & OT \cite{OT} & $34.00_{0.00}$ & $62.45_{0.00}$ & & $81.30_{0.26}$ & $81.30_{0.26}$ & & $43.98_{0.11}$ & $48.86_{0.12}$ & & $86.48_{0.07}$ & $86.69_{0.05}$ \\
        & Panther+Lin. \cite{PANTHER} & $31.80_{0.27}$ & $60.02_{0.32}$ & & $81.34_{0.07}$ & $81.33_{0.07}$ & & $43.97_{0.03}$ & $48.83_{0.03}$ & & $86.61_{0.12}$ & $86.77_{0.10}$ \\
        & Panther+MLP \cite{PANTHER} & $50.86_{0.00}$ & $76.64_{0.00}$ & & $\mathbf{86.89_{0.26}}$ & $\mathbf{86.89_{0.26}}$ & & $\mathbf{51.20_{0.51}}$ & $\mathbf{56.40_{0.40}}$ & & $87.31_{0.18}$ & $88.06_{0.17}$ \\
        \midrule
        \multirow{2}{*}{\rotatebox{90}{\textbf{Ours}}} 
        & \textbf{ProAlign+Lin.} & $\textcolor{blue}{53.07_{0.00}}$ & $\textcolor{blue}{78.83_{0.00}}$ & & $\textcolor{blue}{86.34_{0.11}}$ & $\textcolor{blue}{86.33_{0.11}}$ & & $46.20_{0.05}$ & $51.21_{0.05}$ & & $\mathbf{89.76_{0.16}}$ & $\mathbf{89.81_{0.12}}$ \\
        & \textbf{ProAlign+MLP} & $\mathbf{56.31_{1.66}}$ & $\mathbf{80.87_{1.30}}$ & & $86.18_{0.45}$ & $86.17_{0.46}$ & & $\textcolor{blue}{50.61_{0.77}}$ & $\textcolor{blue}{55.38_{0.60}}$ & & $\textcolor{blue}{89.63_{1.00}}$ & $\textcolor{blue}{89.60_{0.98}}$ \\
        \bottomrule
    \end{tabular}
    }
\end{table*}

\begin{table*}[ht]
    \centering
    \caption{Comparison results on different classification tasks based on PLIP. The best results of supervised baselines are in red. The best results of unsupervised models are in bold, and the second best ones are in blue.}
    \label{plip}
    \resizebox{\linewidth}{!}{
    \begin{tabular}{clcccccccccccccc}
        \toprule
        & \multirow{2}{*}{\textbf{Method}} & \multicolumn{2}{c}{\textbf{CAMELYON+}} & & \multicolumn{2}{c}{\textbf{TCGA-NSCLC}} & & \multicolumn{2}{c}{\textbf{PANDA}} & & \multicolumn{2}{c}{\textbf{CPTAC}} \\
        \cmidrule{3-4} \cmidrule{6-7} \cmidrule{9-10} \cmidrule{12-13} 
        & & \textbf{B acc} & \textbf{F1} & & \textbf{B acc} & \textbf{F1} & & \textbf{B acc} & \textbf{F1} & &  \textbf{B acc} & \textbf{F1}\\
        \midrule
        \multirow{8}{*}{\rotatebox{90}{\textbf{Supervised.}}} 
        & MAXMIL & $48.94_{0.19}$ & $74.30_{0.14}$ & & $80.87_{1.87}$ & $80.86_{1.88}$ & & $47.37_{0.26}$ & $52.65_{0.25}$ & & $81.58_{0.52}$ & $82.39_{0.47}$ \\
        & MEANMIL & $38.06_{0.27}$ & $66.16_{0.24}$ & & $78.17_{0.40}$ & $78.16_{0.40}$ & & $47.37_{0.26}$ & $52.65_{0.25}$ & & $83.18_{0.19}$ & $83.79_{0.21}$ \\
        & ABMIL \cite{ABMIL} & $53.34_{0.85}$ & $78.68_{0.79}$ & & $81.58_{0.57}$ & $81.58_{0.57}$ & & $50.67_{0.42}$ & $56.37_{0.45}$ & & $83.03_{0.54}$ & $83.37_{0.47}$\\
        & DSMIL \cite{DSMIL} & $46.07_{1.17}$ & $72.22_{0.83}$ & & $78.07_{1.06}$ & $78.07_{1.06} $ & & $50.53_{0.45}$ & $56.17_{0.44}$ & & $84.91_{0.27}$ & $85.09_{0.33}$ \\
        & TransMIL \cite{TransMIL} & $\textcolor{red}{61.67_{1.04}}$ & $\textcolor{red}{85.06_{0.72}}$ & & $82.42_{1.13}$ & $82.38_{1.15}$ & & $51.12_{1.76}$ & $56.67_{1.63}$ & & $88.82_{1.18}$ & $89.06_{1.08} $ \\
        & RRTMIL \cite{rrtmil} & $61.31_{1.76}$ & $83.70_{1.21}$ & & $\textcolor{red}{82.86_{1.12}}$ & $\textcolor{red}{82.82_{1.14}}$ & & $53.14_{1.11}$ & $58.76_{1.02}$ & & $88.50_{0.97}$ & $89.00_{0.73}$ \\
        & WiKG \cite{WiKG} & $48.47_{4.21}$ & $74.46_{2.81}$ & & $80.90_{0.85}$ & $80.87_{0.85} $ & & $\textcolor{red}{54.37_{0.95}}$ & $\textcolor{red}{59.66_{0.96}}$ & & $\textcolor{red}{90.85_{0.59}}$ & $\textcolor{red}{90.85_{0.61}}$ \\
        & FRMIL \cite{FRMIL} & $52.13_{9.04}$ & $77.71_{7.44}$ & & $77.95_{1.53}$ & $77.92_{1.53}$ & & $51.26_{1.27}$ & $57.00_{1.11}$ & & $83.03_{1.26}$ & $83.45_{1.15}$ \\
        \midrule
        \multirow{5}{*}{\rotatebox{90}{\textbf{Unsup.}}} 
        & H2T \cite{H2T} & $25.00_{0.00}$ & $50.51_{0.00}$ & & $68.42_{0.14}$ & $68.33_{0.14}$ & & $34.96_{0.07}$ & $38.47_{0.03}$ & & $38.51_{0.06}$ & $44.24_{0.09}$ \\
        & ProtoCount \cite{ProtoCount} & $26.26_{2.53}$ & $31.36_{12.13}$ & & $52.89_{3.18}$ & $48.87_{6.89}$ & & $24.97_{2.02}$ & $26.90_{2.47}$ & & $14.83_{7.81}$ & $11.64_{3.19}$ \\
        & OT \cite{OT} & $24.97_{0.06}$ & $50.48_{0.08}$ & & $66.18_{0.26}$ & $65.35_{0.28}$ & & $44.46_{0.20}$ & $49.56_{0.22}$ & & $73.60_{0.22}$ & $78.56_{0.36}$ \\
        & Panther+Lin. \cite{PANTHER} & $29.80_{0.97}$ & $57.57_{1.26}$ & & $78.57_{0.38}$ & $78.57_{0.38}$ & & $45.06_{0.22}$ & $50.32_{0.16}$ & & $80.39_{0.20}$ & $80.74_{0.15}$ \\
        & Panther+MLP \cite{PANTHER} & $\textcolor{blue}{41.45_{1.12}}$ & $\textcolor{blue}{68.25_{0.88}}$ & & $\mathbf{82.11_{0.26}}$ & $\mathbf{82.11_{0.25}}$ & & $\mathbf{48.88_{0.45}}$ & $\mathbf{54.52_{0.48}}$ & & $\textcolor{blue}{83.01_{0.23}}$ & $\mathbf{83.61_{0.31}}$ \\
        \midrule
        \multirow{2}{*}{\rotatebox{90}{\textbf{Ours}}} 
        & \textbf{ProAlign+Lin.} & $32.64_{0.20}$ & $60.47_{0.19}$ & & $73.63_{0.20}$ & $73.61_{0.19}$ & & $38.30_{0.19}$ & $42.84_{0.22}$ & & $69.70_{0.07}$ & $69.95_{0.09}$ \\
        & \textbf{ProAlign+MLP} & $\mathbf{45.88_{1.40}}$ & $\mathbf{72.52_{1.25}}$ & & $\textcolor{blue}{80.09_{0.76}}$ & $\textcolor{blue}{78.57_{3.60}}$ & & $\textcolor{blue}{46.56_{0.39}}$ & $\textcolor{blue}{51.97_{0.45}}$ & & $\mathbf{83.11_{0.85}}$ & $\textcolor{blue}{83.47_{0.63}}$ \\
        \bottomrule
    \end{tabular}
    }
\end{table*}

We select eight weakly supervised baselines: MAXMIL, MEANMIL, ABMIL\cite{ABMIL}, DSMIL\cite{DSMIL}, TransMIL\cite{TransMIL}, RRTMIL\cite{rrtmil}, WiKG\cite{WiKG}, FRMIL\cite{FRMIL} and four unsupervised baselines: H2T\cite{H2T}, ProtoCount\cite{ProtoCount}, OT\cite{OT}, Panther\cite{PANTHER}, and conduct comparative experiments on four public datasets. The results are shown in Table \ref{conch} and Table \ref{plip}.


As shown in Table \ref{conch}, when using CONCH as encoder, ProAlign demonstrates superior performance compared to other unsupervised models. For instance, in CAMELYON+, ProAlign achieves a balanced accuracy of 56.31\% and a weighted F1 score of 80.87\%, outperforming the next-best-performing model, Panther, by 5.45\% and 4.23\%, respectively. In CPTAC, ProAlign reaches a balanced accuracy of 89.63\% and a weighted F1 score of 89.60\%, surpassing Panther by 2.32\% and 1.54\%, respectively. Additionally, in TCGA-NSCLC and PANDA, ProAlign's overall performance is comparable to Panther's (86.34\% vs. 86.89\% and 50.61\% vs. 51.20\% in balanced accuracy; 86.33\% vs. 86.89\% and 55.38\% vs. 56.40\% in weighted F1 score). However, when using a lightweight linear classifier, ProAlign significantly outperforms Panther. For instance, ProAlign+Lin. achieves balanced accuracies of 86.34\% and 46.20\% in TCGA-NSCLC and PANDA, respectively, exceeding Panther+Lin.'s 81.34\% and 43.67\% by 5\% and 2.03\%, respectively. Lastly, ProAlign exhibits performance comparable to several weakly supervised baselines. In CAMELYON+, ProAlign's balanced accuracy of 56.31\% is 12.95\% higher than DSMIL's 43.36\%. In TCGA-NSCLC, ProAlign's weighted F1 score exceeds five of eight weakly supervised baselines.


As indicated in Table \ref{plip}, when using PLIP as encoder, ProAlign is the best or second-best unsupervised model across all four tasks, demonstrating competitive performance. For example, in CAMELYON+, ProAlign achieves a balanced accuracy of 45.88\% and a weighted F1 score of 72.52\%, outperforming the next-best-performing model, Panther, by 4.43\% and 4.27\%, respectively. In CPTAC, ProAlign's balanced accuracy of 83.11\% leads Panther's 83.01\%. Overall, compared to CONCH, all models exhibit varying degrees of performance decline when using PLIP as feature extractor, with a more significant impact on unsupervised models. In this context, ProAlign still demonstrates superior performance over some weakly supervised models. For instance, in TCGA-NSCLC, ProAlign achieves a balanced accuracy of 80.09\%, surpassing FRMIL's 77.95\% by 2.04\%.

\begin{figure}[ht]
  \centering
  \includegraphics[width= \linewidth]{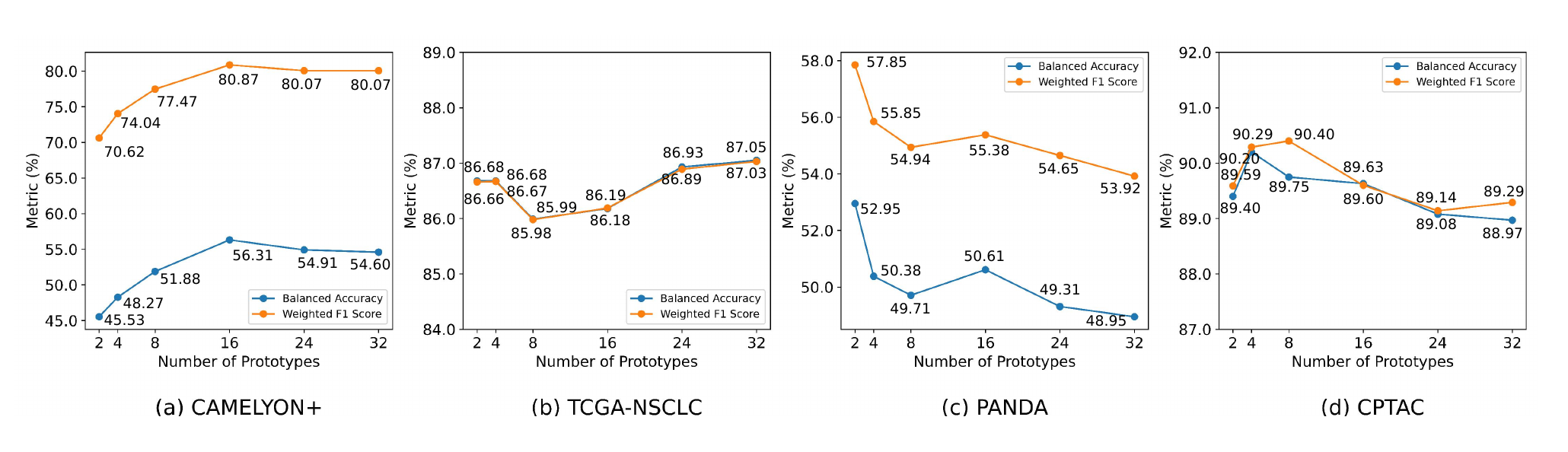}
   \caption{Hyperparameter study of the number of prototypes}
   \label{as}
\end{figure}

\subsection{Effectiveness of the number of prototypes}

To investigate the impact of prototype quantity on model performance, we employ LLM to condense and expand the original 16 prototype descriptions, creating additional sets of 2, 4, 8, 24, and 32 prototypes. Textual embeddings are extracted using CONCH, and experiments are conducted across four publicly available datasets, with results presented in Fig. \ref{as}. In general, model performance exhibits data-dependent sensitivity to the number of prototypes. In CAMELYON+, as the number of prototypes increases, the performance of the model initially rises and then stabilizes. Similarly, in TCGA-NSCLC, despite some fluctuations, there is an overall upward trend in performance with increasing prototype numbers. In contrast, in PANDA, model performance declines as the number of prototypes increases, which is related to the fact that each WSI in the PANDA dataset contains only a few dozen patches on average. For CPTAC, model performance fluctuates within a 1\% range as the number of prototypes increases, with minimal overall change. 

\subsection{Visualization}
\begin{figure}[t]
  \centering
  \includegraphics[width= \linewidth]{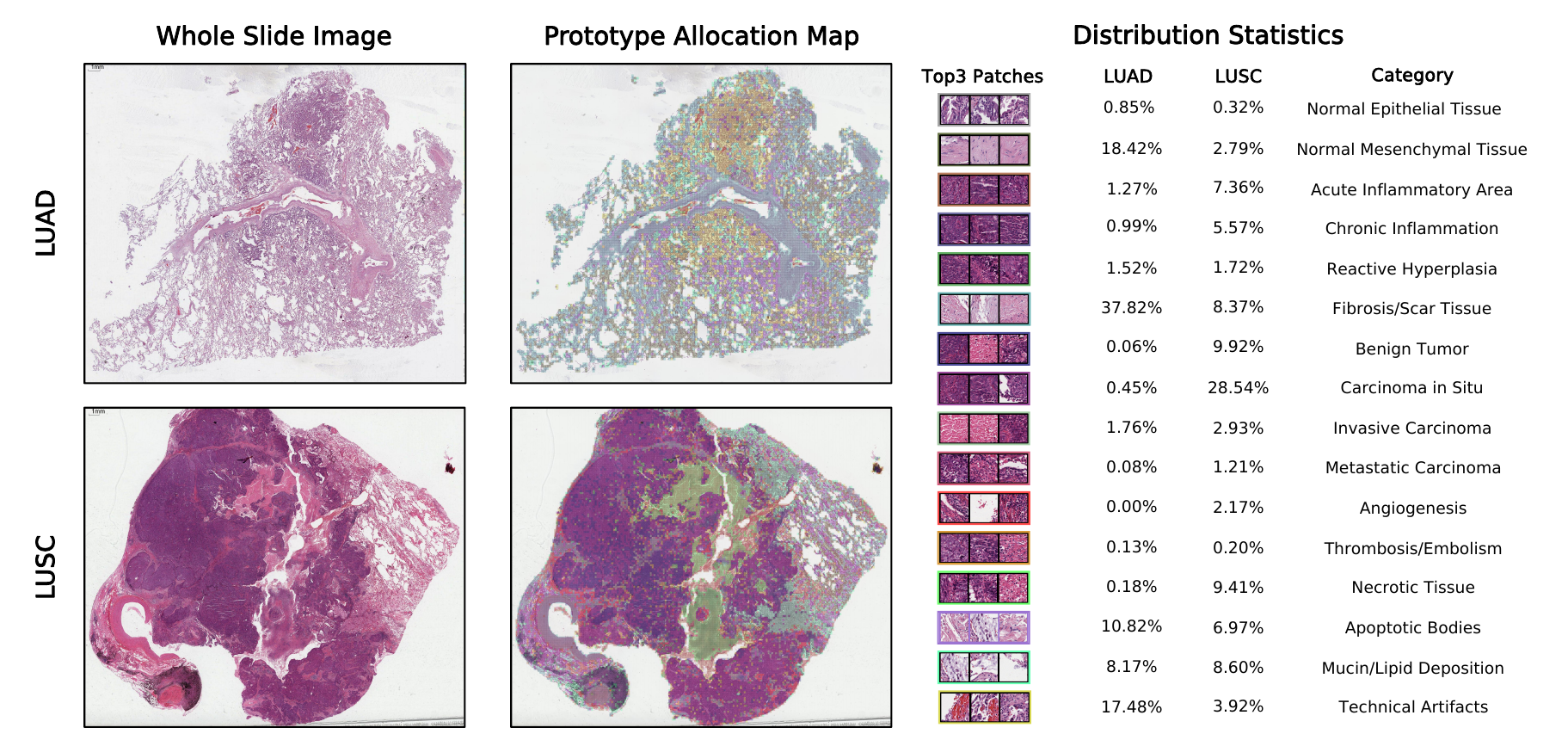}
   \caption{Prototype allocation map visualization, including a LUAD slide and a LUSC slide. Distributions statics shows the 3 most similar patch images to each prototype, the proportion of each prototype in the LUAD slide and the LUSC slide, and the specific name of each prototype}
   \label{visual}
\end{figure}

To assess the rationality of our prototype allocation, we visualize prototype allocation maps compiled with distribution data for two categories of slides, lung adenocarcinoma (LUAD) and lung squamous cell carcinoma (LUSC), from TCGA-NSCLC, as shown in Fig. \ref{visual}. The allocation of prototypes on both slides is consistent with their characteristic biological features. For example, "Fibrosis/Scar Tissue" accounts for 37.82\%, indicating a significant presence of fibrotic responses, which are commonly associated with chronic inflammation and tissue repair in tumors. This high proportion is consistent with the typical fibrotic characteristics observed in LUAD slides, as tumor growth and metastasis are frequently accompanied by tissue remodeling and fibrosis. Additionally, the high proportion of "Carcinoma in Situ" at 28.54\% reflects a significant presence of localized tumor infiltration during the early stages of LUSC. LUSC typically originates from the epithelial cells of the airways, and slides from this category often show precancerous lesions such as carcinoma in situ.

\section{Conclusion}
In this paper, we propose ProAlign, a cross-modal unsupervised slide representation learning framework. ProAlign optimizes the prototype construction process by incorporating textual modality information from prototype descriptions through LLM prompting. And then aggregates the prototypes using a parameter-free attention mechanism to obtain slide representations. Extensive experiments on multiple public datasets demonstrate that ProAlign outperforms existing unsupervised baselines and is on par with several strong weakly supervised baselines.
\bibliographystyle{splncs04}
\bibliography{Paper-2626}
\end{document}